\documentclass[conference]{IEEEtran}
\usepackage{times}

\usepackage[numbers]{natbib}
\usepackage{multicol}
\usepackage[bookmarks=true]{hyperref}

\usepackage{graphics} 
\usepackage{epsfig} 
\usepackage{amsmath} 
\usepackage{amssymb}  
\usepackage{comment}
\usepackage{amsmath,amssymb} 
\usepackage{color}
\usepackage{multirow}
\usepackage{url}
\usepackage{hyperref}
\usepackage{graphicx}
\usepackage{subfigure} 
\usepackage{float}

\usepackage{lipsum}

\usepackage{hyperref}
\usepackage[hyphenbreaks]{breakurl}

\newcommand\blfootnote[1]{%
  \begingroup
  \renewcommand\thefootnote{}\footnote{#1}%
  \addtocounter{footnote}{-1}%
  \endgroup
}

\newcommand{\ie}{\textit{i}.\textit{e}., }
\newcommand{\eg}{\textit{e}.\textit{g}. }

\begin{document}

\title{RFUniverse: A Multiphysics Simulation Platform for Embodied AI}


\author{\authorblockN{Haoyuan Fu*}
\authorblockA{Shanghai Jiao Tong University\\
simon-fuhaoyuan@sjtu.edu.cn}
\and
\authorblockN{Wenqiang Xu*}
\authorblockA{Shanghai Jiao Tong University\\
vinjohn@sjtu.edu.cn}
\and
\authorblockN{Ruolin Ye*}
\authorblockA{Cornell University\\
ry273@cornell.edu}
\and
\authorblockN{Han Xue}
\authorblockA{Shanghai Jiao Tong University\\
xiaoxiaoxh@sjtu.edu.cn}
\and
\authorblockN{Zhenjun Yu}
\authorblockA{Shanghai Jiao Tong University\\
jeffson-yu@sjtu.edu.cn}
\and
\authorblockN{Tutian Tang}
\authorblockA{Shanghai Jiao Tong University\\
tttang@sjtu.edu.cn}
\and
\authorblockN{Yutong Li}
\authorblockA{Shanghai Jiao Tong University\\
davidliyutong@sjtu.edu.cn}
\and
\authorblockN{Wenxin Du}
\authorblockA{Shanghai Jiao Tong University\\
mnkmYuki@sjtu.edu.cn}
\and
\authorblockN{Jieyi Zhang}
\authorblockA{Shanghai Jiao Tong University\\
yi\_eagle@sjtu.edu.cn}
\and
\authorblockN{Cewu Lu}
\authorblockA{Shanghai Jiao Tong University\\
lucewu@sjtu.edu.cn}
}


%

\maketitle

\begin{abstract}
Multiphysics phenomena, the coupling effects involving different aspects of physics laws, are pervasive in the real world and can often be encountered when performing everyday household tasks. 
Intelligent agents which seek to assist or replace human laborers will need to learn to cope with such phenomena in household task settings. To equip the agents with such kind of abilities, the research community needs a simulation environment, which will have the capability to serve as the testbed for the training process of these intelligent agents, to have the ability to support multiphysics coupling effects. 

Though many mature simulation software for multiphysics simulation have been adopted in industrial production, such techniques have not been applied to robot learning or embodied AI research. To bridge the gap, we propose a novel simulation environment named \textbf{RFUniverse}. 
This simulator can not only compute rigid and multi-body dynamics, but also multiphysics coupling effects commonly observed in daily life, such as air-solid interaction, fluid-solid interaction, and heat transfer. 

Because of the unique multiphysics capacities of this simulator, we can benchmark tasks that involve complex dynamics due to multiphysics coupling effects in a simulation environment before deploying to the real world. RFUniverse provides multiple interfaces to let the users interact with the virtual world in various ways, which is helpful and essential for learning, planning, and control. We benchmark three tasks with reinforcement learning, including food cutting, water pushing, and towel catching. We also evaluate butter pushing with a classic planning-control paradigm. This simulator offers an enhancement of physics simulation in terms of the computation of multiphysics coupling effects.
The simulation environment, videos, and other supplementary materials can be viewed on the website: \url{https://sites.google.com/view/rfuniverse}.
\end{abstract}

\blfootnote{* Denotes equal contribution. Cewu Lu is the corresponding author, and member of Qing Yuan Research Institute and MoE Key Lab of Artificial Intelligence, AI Institute, Shanghai Jiao Tong University, China. Ruolin Ye was with Shanghai Jiao Tong University when work was done.}

\IEEEpeerreviewmaketitle

\begin{figure}[ht!]
    \centering
    \hspace{-3mm}
    \subfigure[Navigation.]{
        \begin{minipage}[t]{0.3\linewidth}
            \includegraphics[width=1\linewidth]{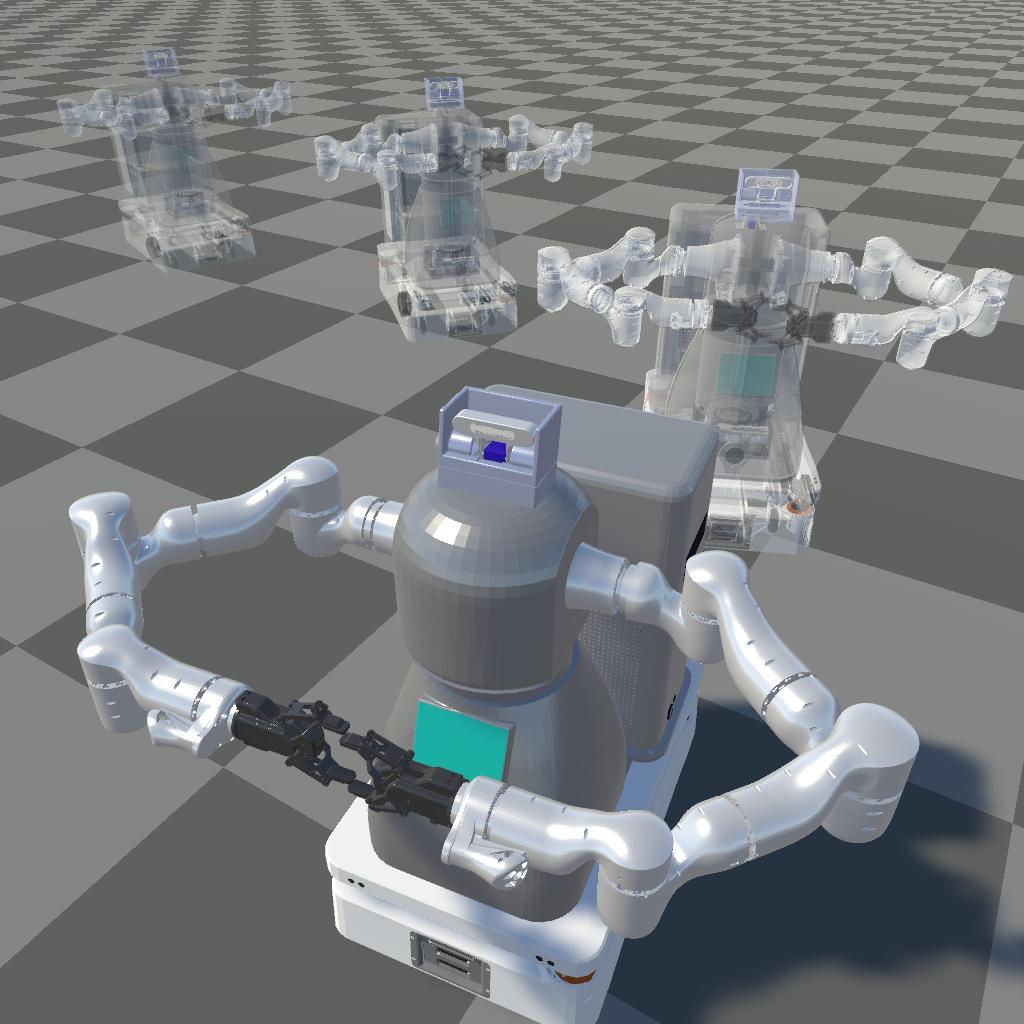}
            \vspace{-4mm}
            \label{}
        \end{minipage}
    }
    \hspace{-3mm}
    \subfigure[Pick and place.]{
        \begin{minipage}[t]{0.3\linewidth}
            \includegraphics[width=1\linewidth]{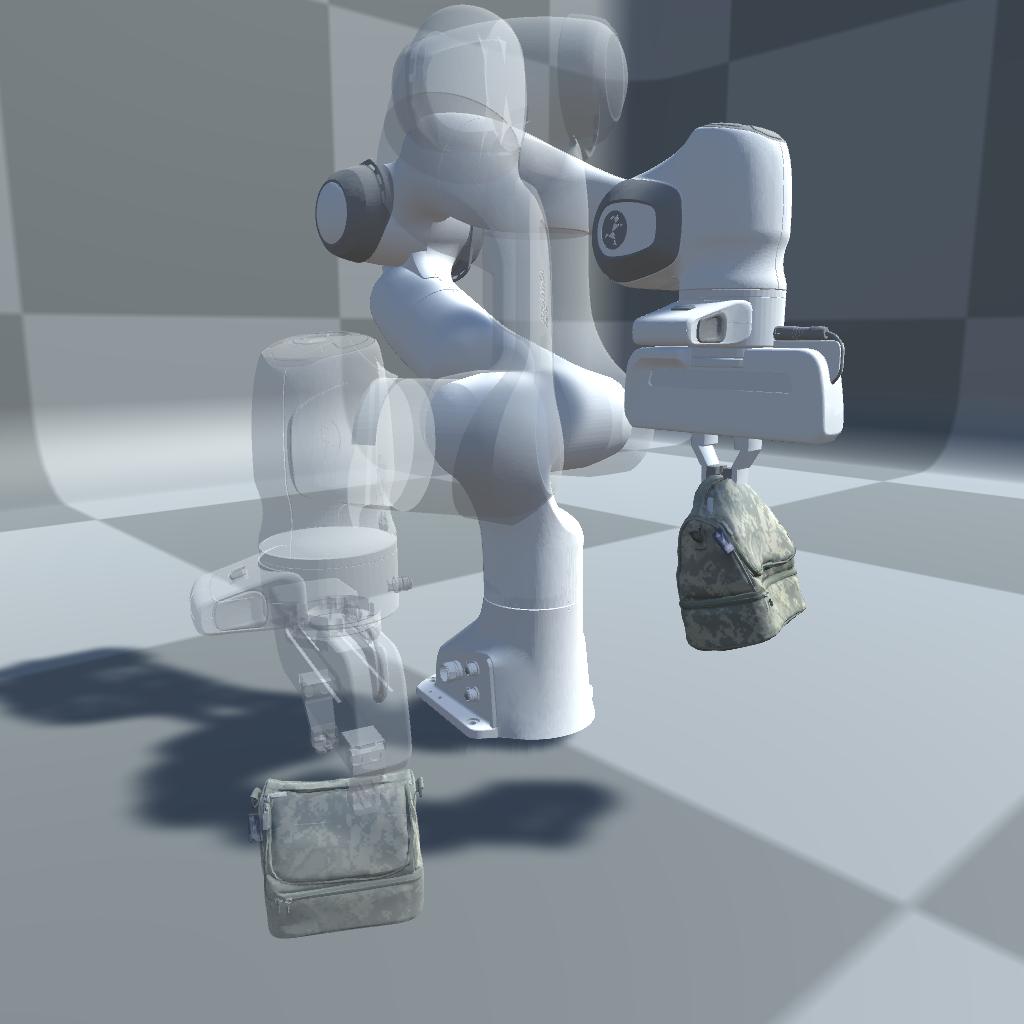}
            \vspace{-4mm}
            \label{}
        \end{minipage}
    }
    \hspace{-3mm}
    \vspace{-2mm}
    \subfigure[Cutting.]{
        \begin{minipage}[t]{0.3\linewidth}
            \includegraphics[width=1\linewidth]{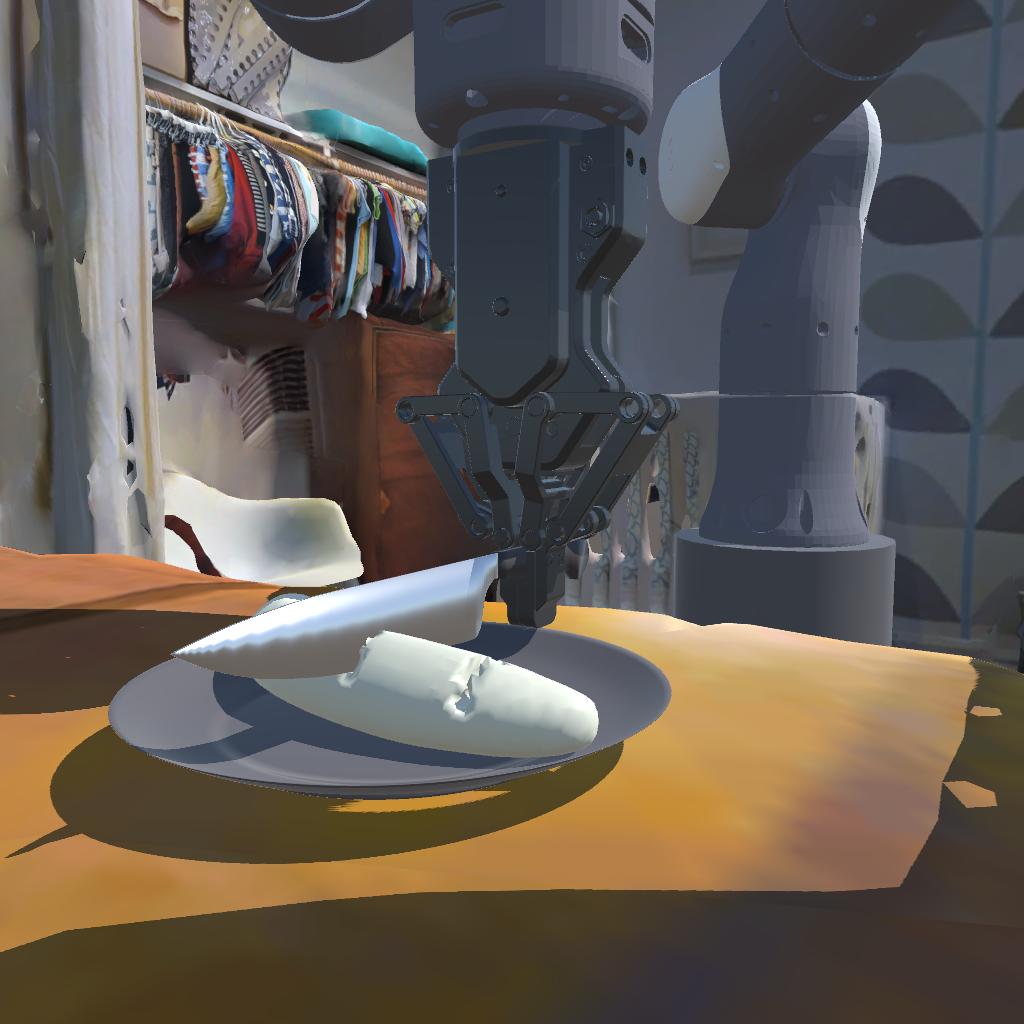}
            \vspace{-4mm}
            \label{}
        \end{minipage}
    }
    \subfigure[Water pushing.]{
        \begin{minipage}[t]{0.3\linewidth}
            \includegraphics[width=1\linewidth]{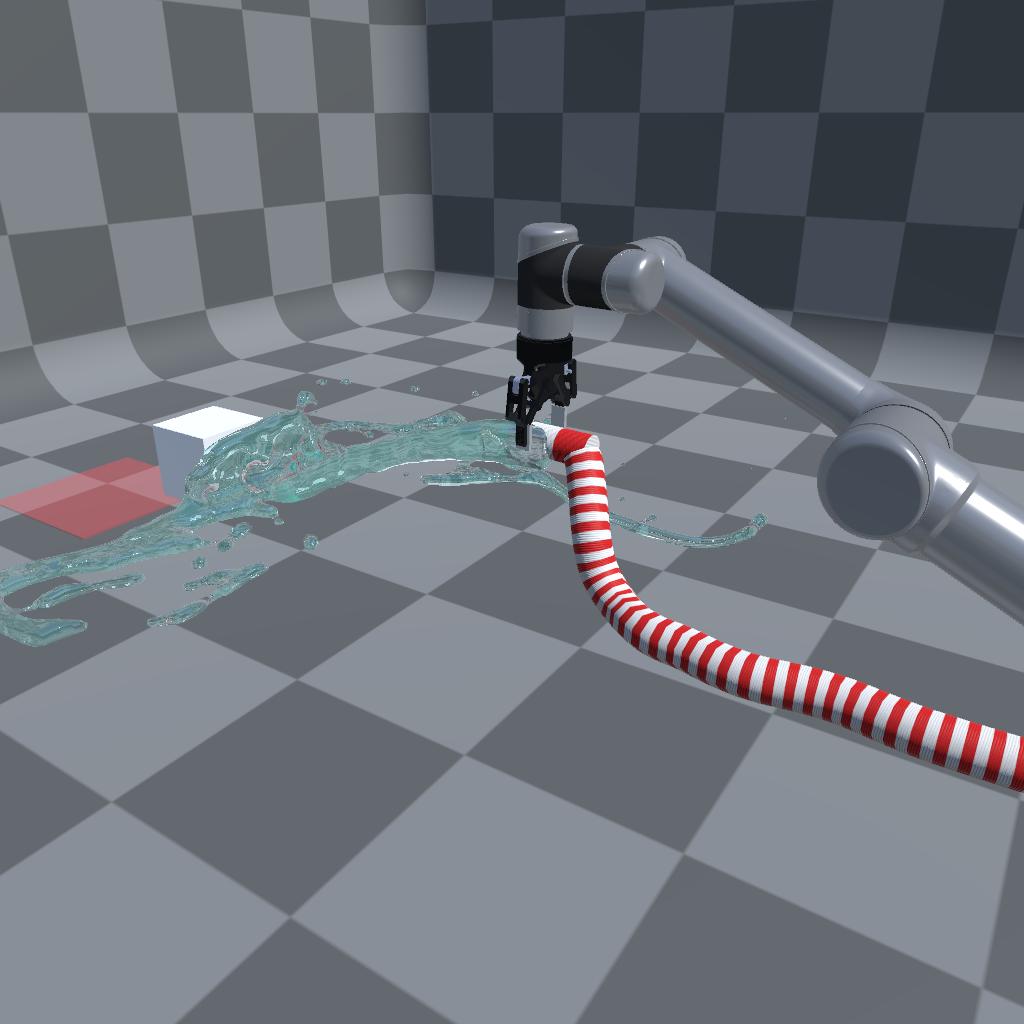}
            \vspace{-4mm}
            \label{}
        \end{minipage}
    }
    \hspace{-3mm}
    \subfigure[Towel catching.]{
        \begin{minipage}[t]{0.3\linewidth}
            \includegraphics[width=1\linewidth]{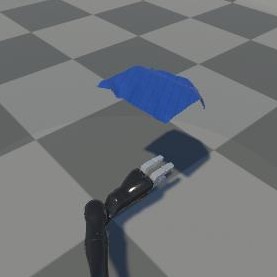}
            \vspace{-4mm}
            \label{}
        \end{minipage}
    }
    \hspace{-3mm}
    \subfigure[Butter pushing.]{
        \begin{minipage}[t]{0.3\linewidth}
            \includegraphics[width=1\linewidth]{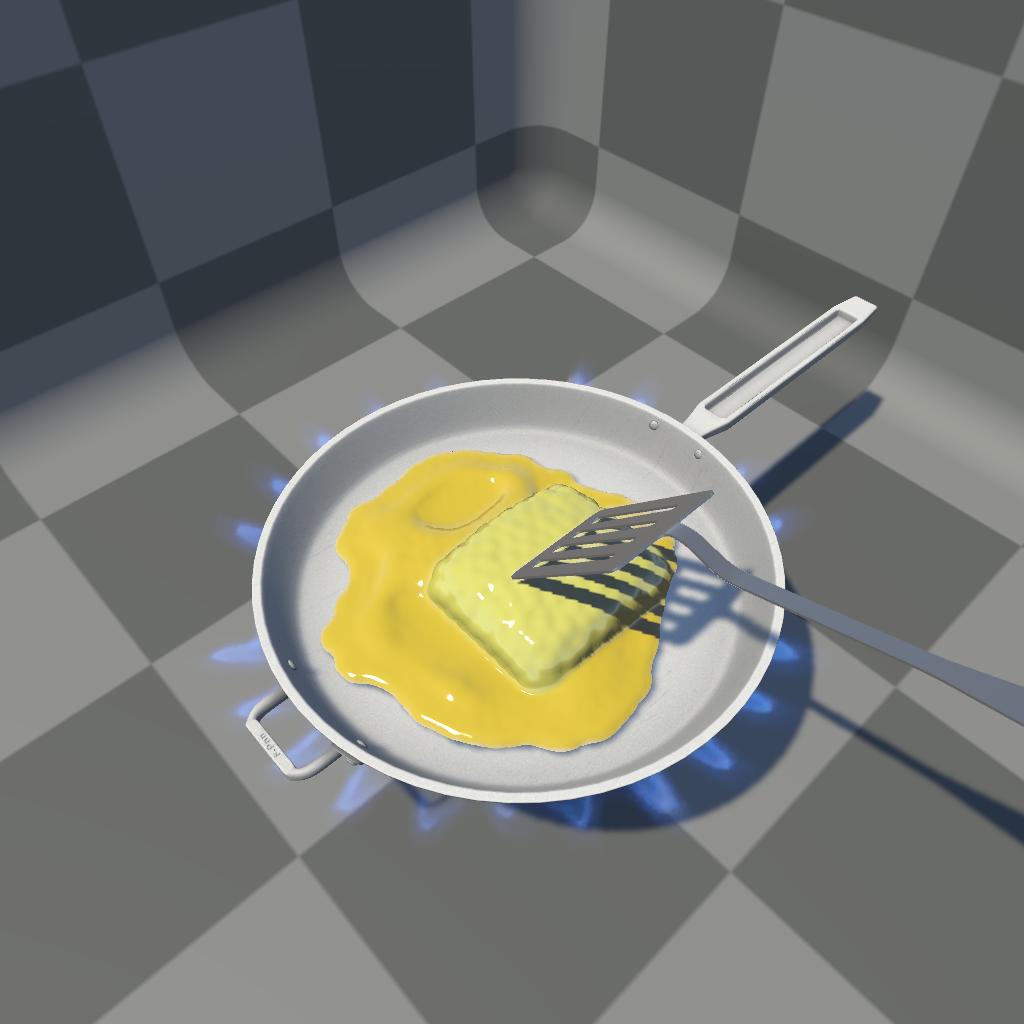}
            \vspace{-4mm}
            \label{}
        \end{minipage}
    }
    \caption{Tasks in RFUniverse: Standard tasks such as (a) Navigation (b) Pick and place, and tasks involving multiphysics interaction such as  (c) Physics-based cutting; (d) Water pushing; (e) Towel catching; (f) Butter pushing. }
    \label{fig:fig1}
\end{figure}
\section{INTRODUCTION}
Physics laws govern the real world and therefore generate various object dynamics.
Multiphysics phenomena can happen when objects in different states interact with each other. 
Humans can encounter many these common phenomena when conducting everyday household tasks. 
For instance, a piece of towel dropping in the air can easily drift its path due to \textit{aerodynamics}. The water from a faucet can flush away the food waste and dirts through \textit{fluid-solid interaction}. 
The \textit{heating} process can soften the butter and transfer its state from solid to liquid.
An intelligent agent which learns to perform household tasks will need to learn to react with object state changes due to these common multiphysics coupling effects.

However, though many simulation environments \cite{ai2thor,gibson,home,virtualhome,alfred,habitat,igibson0.5,igibson1,igibson2,manipulathor, robothor, vrkitchen, sapien, rcare} have been developed for embodied AI research in recent years,  
they usually focus on tasks which only require the basic rigid and multi-body dynamics simulation, such as abstract decision-making or simple manipulation. 
Almost none of them considers the object dynamics under coupled physics effects. 
A recent simulation environment OmniGibson \cite{behavior1k} supports over 2K household tasks, which covers the largest scope of household tasks by far. However, even though it contains tasks that involve complex physics coupling effect, like water pouring and gas ignition, the underlying multiphysics interaction is still simplified or ignored in this environment. These kinds of simplification limit the task scope of embodied AI research and limits the sim-to-real transfer ability of the algorithms developed in simulation.

To bridge the gap and extend the task scope of embodied AI research, we present a novel simulation environment named \textbf{RFUniverse}. Aside from common rigid and multi-body dynamics simulation, RFUniverse integrates three kinds of multiphysics coupling effects into the simulation environment, namely air-solid interaction (aerodynamics), fluid-solid interaction (hydrodynamics), and heat transfer (thermodynamics). Without a doubt, in nature, different physics laws have more than these three ways of coupling to influence the world. However, it is not practical to integrate all kinds of multiphysics effects into the simulation while maintaining a real-time performance which is essential for robot learning and control. Therefore, we choose to implement these three multiphysics effects which are the most commonly encountered in our daily life and therefore will have the potential to better support common household tasks. With the underlying support of multiphysics, there will be chances for the research community to evaluate the execution process for advanced physics-based tasks. We focus on four key tasks in this paper, including physics-based \textbf{food cutting}, \textbf{water pushing}, \textbf{towel catching}, and \textbf{butter pushing}.


\begin{table*}[t!]
\caption{}\label{tab:big_comp}
\begin{center}
\begin{tabular}{|c|p{4cm}|c|c|c|c|c|}
\hline
Environment & Physics & Learning & Rendering& Integrated Planner & VR & ROS\\
\hline
COSMOL \cite{cosmol} &aero, hydro, thermo, multi-body &/ & /&/ & / &/\\
Abaqus \cite{abaqus}&aero, hydro, thermo, multi-body &/ & / &/ & / & /\\
Chrono \cite{chrono}& hydro, multi-body& /& / &/& / & /\\
AGX Dynamics \cite{agx}& aero, hydro, multi-body &\checkmark & photo-&\checkmark&/&/\\
\hline
HoME \cite{home} & multi-body & \checkmark & photo- &  \checkmark & / & /\\
AI2THOR series \cite{ai2thor, robothor, manipulathor} & multi-body & \checkmark & photo- &  / & / &/\\
Habitat \cite{habitat} &multi-body  & \checkmark & photo- & / & / & / \\
SAPIEN \cite{sapien} & multi-body& \checkmark & physics- &\checkmark & \checkmark & / \\
VirtualHome \cite{virtualhome} &multi-body & \checkmark & photo-& / & / & /\\
VRKitchen \cite{vrkitchen} & multi-body & \checkmark & photo- & / &/ &\checkmark\\
ThreeDWorld \cite{threedworld,threedworldChallenge} &multi-body & \checkmark & photo- & / & \checkmark & /\\
Gibson series \cite{gibson,igibson0.5,igibson1,igibson2,behavior1k} & multi-body & \checkmark & physics-& $\checkmark$ & \checkmark & \checkmark \\
Isaac Gym \cite{isaacgym} & multi-body & \checkmark & physics-& / & / & / \\
\hline
RFUniverse & aero, hydro, thermo, multi-body & \checkmark &physics- &  \checkmark & \checkmark  & \checkmark\\\hline
\end{tabular}
\end{center}
\textbf{Comparison with different simulation environments. As some environments are developed in a long run as different versions, we summarize them into a series and show the features based on the latest version.}\\
\textbf{Physics:} ``aero'' for aerodynamics, ``hydro'' for hydrodynamics, ``thermo'' for thermodynamics, ``multi-body'' for multi-body dynamics. We check if it has at least one function that involves such dynamics. To note, some simulators exhibit water rendering, but such fluid cannot provide interaction force. \\ 
\textbf{Learning:} whether support reinforcement learning or support to produce a visual dataset for perception tasks.\\
\textbf{Rendering:} ``photo-'' for photorealistic rendering effect, ``physics-'' for physics-based rendering effect (at least ray-tracing enabled).\\
\end{table*}

Admittedly, a high-accuracy calculation of multiphysics interaction~\cite{cosmol,abaqus} requires considerable computational resources so that is difficult to run in real time, making it not suitable for the real-time requirement from the robotics community. To balance usability and accuracy, we adopt solutions with simplification on the physics side with a decent computation speed while maintaining fidelity to real-world dynamics. 
To prove the fidelity, we conduct parallel experiments to compare the multiphysics effects in the virtual world with the ones in the real world. For example, with a glass of water, we observe its weight and volume and find out it is the same in the real world. We fix a towel and blow it with the wind, finding the virtual world and the real one are quite aligned in terms of the movement of the towel. We heat butter and see it melting down similarly in simulation and the real world.
We hope our simulation environment with verified fidelity can serve as a platform for multiphysics-based robot learning, and facilitate the integration of advanced simulation techniques into embodied AI research. 

To help utilize the simulation environment, in addition to the multiphysics simulation, RFUniverse also provides full functionalities to support task simulation and learning: physics-based rendering, multi-modal sensing, Python APIs, and a gym-like wrapper for reinforcement learning. Besides, we provide a ROS-free version of MoveIt~\cite{moveit}, RFMove \cite{rfmove}, as a lightweight planner, and natively integrate it into RFUniverse for full-body movement planning. In addition, we also provide ROS interface to enlarge the ecosystem and expand the scope of potential users. We also provide a VR interface to extend the interactive ability from the real to the simulated environment.

We present a set of standard tasks trained with reinforcement learning to showcase the usage of RFUniverse to robot learning in supplementary materials, such as navigation in the setting of multi-agent collaboration, locomotion in the setting of goal reaching, and a few manipulation tasks (\eg fruit picking, cloth folding, sponge wiping). 
In the main paper, we would like to highlight and focus more on introducing and explaining the multiphysics-based system, which is seldom considered by previous simulation environments. We evaluate food cutting, water pushing, towel catching with reinforcement learning, and butter pushing with classic planning control, to demonstrate the usage of this multiphysics-based system. We also perform experiments for visual pre-trained encoders for reaching and cabinet closing.

We summarize our contributions as follows:
\begin{itemize}
    \item We propose a multiphysics-based simulation environment RFUniverse for embodied AI research. RFUniverse provides a client-server communication framework based on gRPC, which can enable full functionality control of Unity with multiple interfaces. It provides physics-based rendering and multi-modal sensing, enabling the agent to perceive the physics information in the virtual world.
    \item With consideration of the balance between fidelity and computational cost, RFUniverse takes the first step towards building a simulation platform for manipulation tasks involving multiphysics dynamics in robot manipulation, including structural mechanics, aerodynamics, hydrodynamics, and thermodynamics. 
    \item We conduct extensive parallel experiments and prove the fidelity of the real-time simulation by comparing them with the real world in the scope of tasks involving multiphysics phenomena, and verify the fidelity of the physics computation in the simulated environment. 
    \item With RFUniverse, we extend the task scope to multiphysics-based tasks, which have rarely been explored before. With the extended task scope, we are able to benchmark three tasks with reinforcement learning, namely physics-based cutting, water pushing, and towel catching; we benchmark one task with planning and control for butter pushing.
\end{itemize}

\section{RELATED WORKS}
\subsection{Multiphysics Simulation Software} 
In the field of modern engineering and science research, the problems left to solve usually require solutions that span a multitude of physical phenomena. There are a lot of commercial (\eg COSMOL \cite{cosmol}, Abaqus \cite{abaqus}) or open-source (\eg MOOSE \cite{moose}) softwares designed for such demands. However, due to the requirement for high-accuracy, the algorithms behind these softwares prefer to take a lot of time and computational resources to process, which make them hard to be integrated into a learning pipeline, where the simulation results should be produced on-the-fly. 

Among these softwares, only a few provide interfaces for robotics, namely Project Chrono \cite{chrono}, and AGX dynamics \cite{agx}. And only AGX dynamics supports real-time simulation. From the multiphysics perspective, AGX dynamics supports fluid mechanics and aerodynamics. The solid mechanics in AGX dynamics are limited to the APIs it provides.

In our solution, we manage to integrate different physics backends together, namely, PhysX \cite{PhysX}, SOFA \cite{sofa}, Obi \cite{obi}, Zibra \cite{zibra}. These physics backends do not provide robot interfaces on their own, but they have specialties to simulate different physics phenomena. 

\subsection{Physics-based Simulation in Embodied AI}
In the past years, we have witnessed a rapid growth of simulation environments for embodied AI or robot learning. Most of the simulation environments adopt one physics engine for physics simulation, such as PhysX \cite{PhysX}, MuJoCo \cite{mujoco}, Bullet \cite{pybullet}, Flex \cite{flex}, ODE \cite{ode} and so on. Most of these physics engines can provide high-fidelity rigid and multi-body dynamics simulation. However, not a single physics engine can support all the multiphysics simulations. The limitation of simulation also setback the development of robot learning on some advanced manipulation tasks such as deformable object manipulation, arbitrary object cutting (i.e. not slicing the object in advance), and those involving object state change due to multiphysics interaction.
As a result, though we already have many simulation environments available \cite{ai2thor,gibson,home,virtualhome,alfred,habitat,igibson0.5,igibson1,igibson2,manipulathor, robothor, vrkitchen, sapien, rcare}, among them, a most recent simulation environment OmniGibson \cite{behavior1k} comprises of the most comprehensive household tasks so far. They all fail to consider the multiphysics tasks.

A detailed comparison between RFUniverse and previous environments can be referred to in Table \ref{tab:big_comp}.
\begin{figure}[th!]
  \centering
  \includegraphics[width=0.45\textwidth]{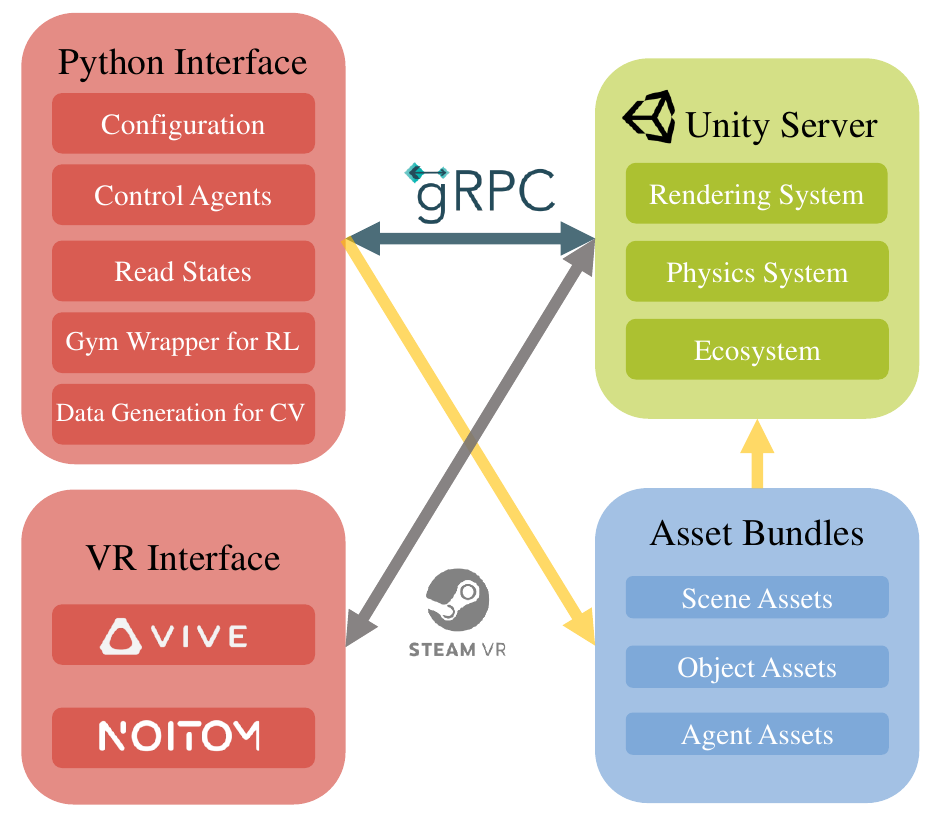}
  \caption{The framework of RFUniverse.}
  \label{fig:framework}
\end{figure}

\section{RFUniverse Simulation Environment}
\textbf{RFUniverse} is a simulation environment that provides high-fidelity physics simulation for multiphysics interaction. Based on a server-client interface, it supports multiple interfaces to interact with the Unity server, which integrates different physics backends for various object types, thus enabling the simulation of multiphysics coupling effects.
\subsection{Communication Framework \& Interfaces}\label{sec:python}
\begin{figure}[th!]
   \centering
  \includegraphics[width=0.41\textwidth]{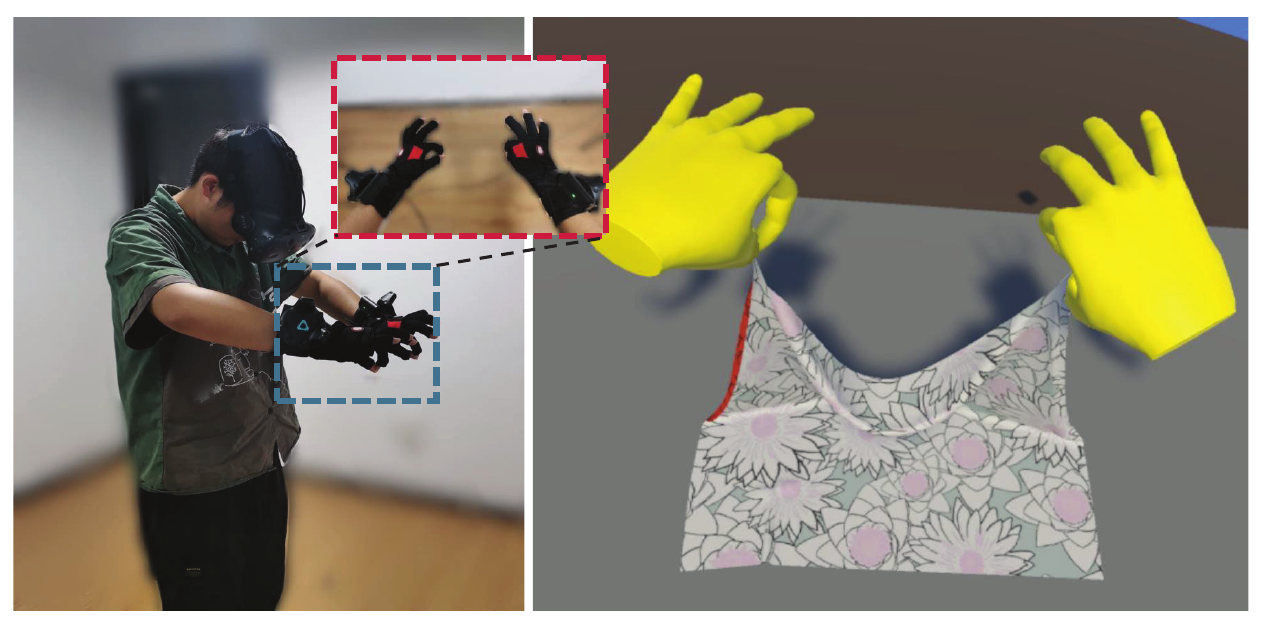}
   \caption{The VR interface with HTC Vive and Noitom Glove.}
   \label{fig:vr_setup}
   \end{figure}

\textbf{RFUniverse} adopts a server-client framework based on gRPC, enabling Python and VR interfaces to communicate with the Unity backend. We show its structure in Fig. \ref{fig:framework}. This communication framework supports multiple languages and OS platforms. We adopt Python as the interfacing language, which is widely used in learning frameworks~\cite{pytorch, mlagents}, due to its simplicity and ecosystem.
Different VR devices can be connected to RFUniverse through SteamVR. We demonstrate the VR interface with the HTC Vive headset and Noitom Hi5 glove in Fig. \ref{fig:vr_setup}.

\subsection{Multiphysics Interaction}
Object state in daily life is governed by different physics laws. Conventionally, these physics phenomena are studied by different branches of mechanics, such as solid mechanics, fluid mechanics (hydrodynamics), aerodynamics, and thermodynamics. 
Despite the fact there are mature commercial softwares for precise computation of these mechanics during multiphysics interaction, they usually cost a decent amount of computational resources and time, making it hard for them to satisfy the need for real-time computation from the robotics side, and therefore seldom considered in embodied AI. To take the first step towards building a simulation environment for the manipulation tasks involving multiphysics dynamics,  we adopt the solvers that balance physics fidelity and computational efficiency and integrate them in Unity. 

In Unity, we borrow the wisdom from modern physics engines including~\cite{sofa} and~\cite{zibra} which support mesh-based and particle-based solvers. To be specific, for solid mechanics, we model the computation process with finite element mechanics with different constitutive models. By treating the object as a volume instead of a 2-manifold surface, it can support cutting the object from arbitrary positions in a physics realistic way. For hydrodynamics, aerodynamics, and thermodynamics, we leverage particle-based methods to simulate the diffusion process of fluid, air, and heat.


\subsection{Physics-based Rendering System}
\begin{figure}[th!]
  \centering
  \includegraphics[width=0.45\textwidth]{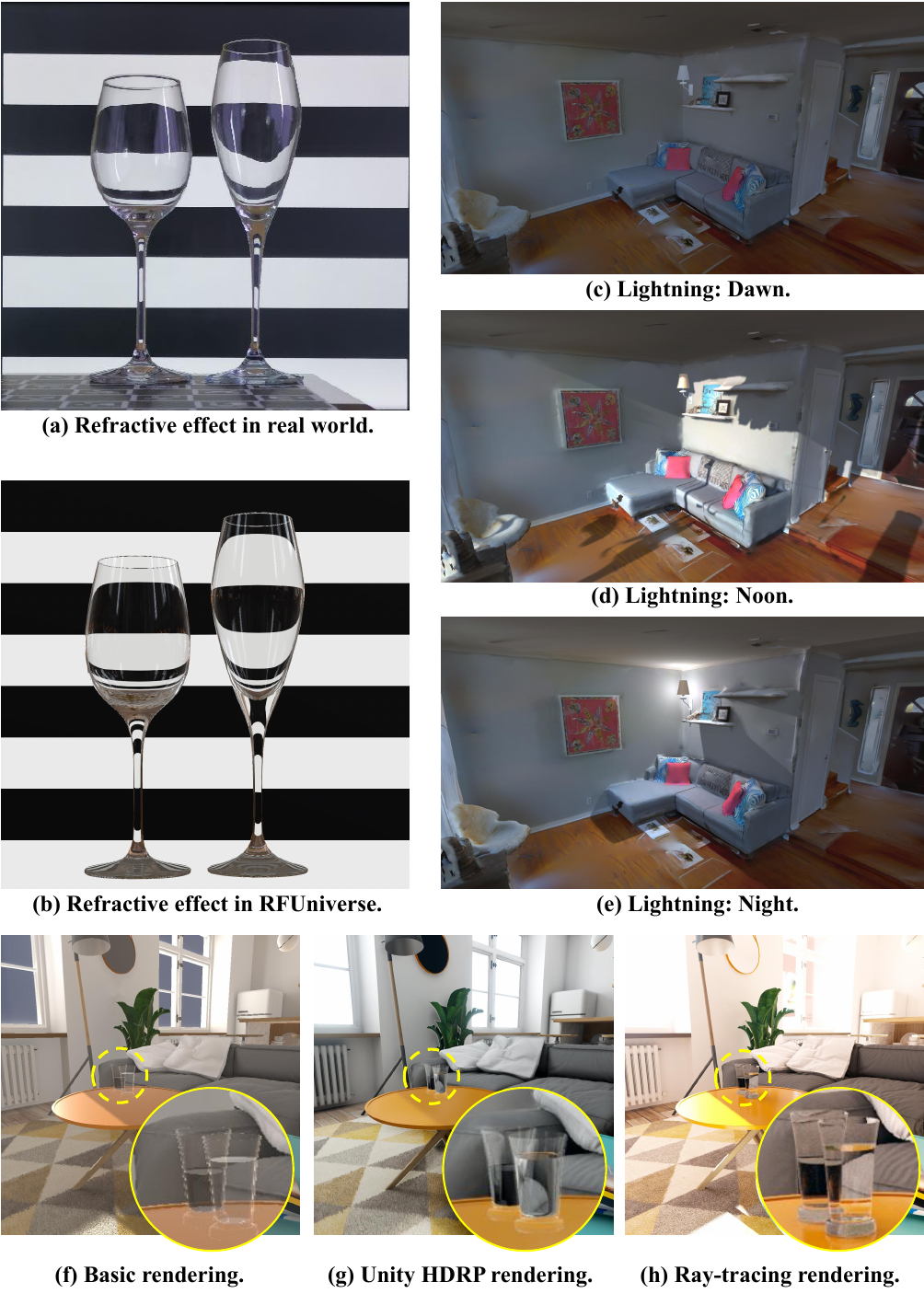}
  \caption{\textbf{(a)(b):} Refractive effect in real world and in RFUniverse. Two wine glasses with gray code behind them. Here we mainly compare the refraction effects because it can reveal the physics fidelity of the rendering system; \textbf{(c)-(e):} different indoor lighting conditions; \textbf{(f)-(h):} Liquid in refraction rendering effects under basic rendering, 'photorealistic' rendering claimed by other simulation environments, and ray-tracing (physics-based) rendering.}
  \label{fig:render}
\end{figure}

Visual input is one of the important modalities for agents to perceive the world. Simulating a high-fidelity physics-based rendering system can mitigate the sim-to-real gap when deploying the algorithms from simulation to the real world. Though many previous simulation environments claim their simulators can provide ``photorealistic'' rendering effects, we realize such photorealism might be unfaithful in physics. As shown in Fig. \ref{fig:render}, a ``photorealistic'' rendering pipeline cannot produce correct refraction effects for a glass and the liquid in a glass. In comparison, RFUniverse supports ray-tracing techniques for rendering and therefore is capable of handling complex optical effects in real-time. It can be rendered at 55 fps on an Nvidia RTX 3090 graphics card. 


\begin{figure*}[h!]
    \centering
    \subfigure[Visual sensor.]{
        \begin{minipage}[t]{0.45\linewidth}
            \includegraphics[width=1\linewidth]{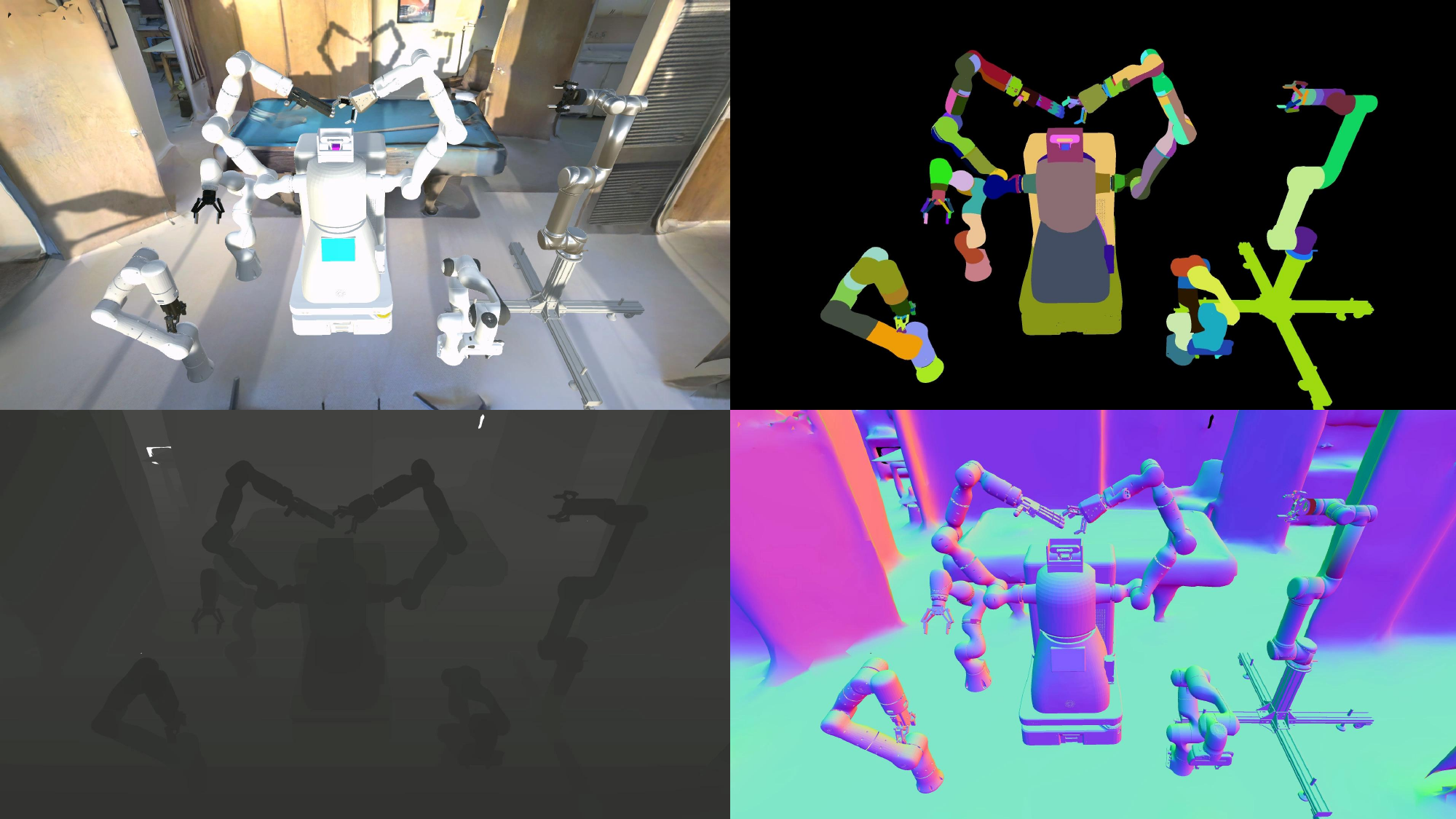}
        \end{minipage}
    }
    \subfigure[IR-based depth sensor.]{
        \begin{minipage}[t]{0.45\linewidth}
            \includegraphics[width=1\linewidth]{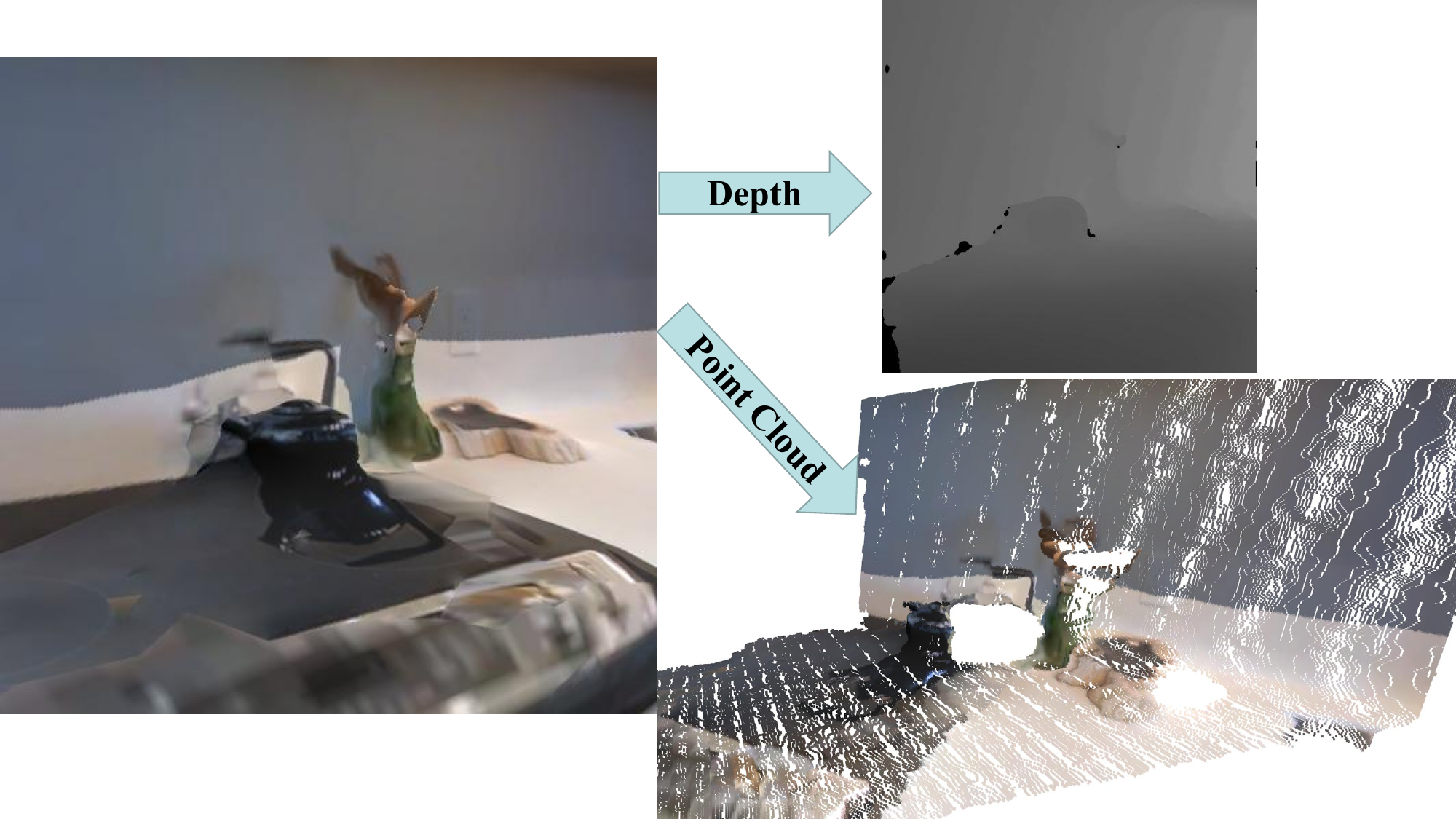}
        \end{minipage}
    }
    \subfigure[Tactile sensor.]{
        \begin{minipage}[t]{0.45\linewidth}
            \includegraphics[width=1\linewidth]{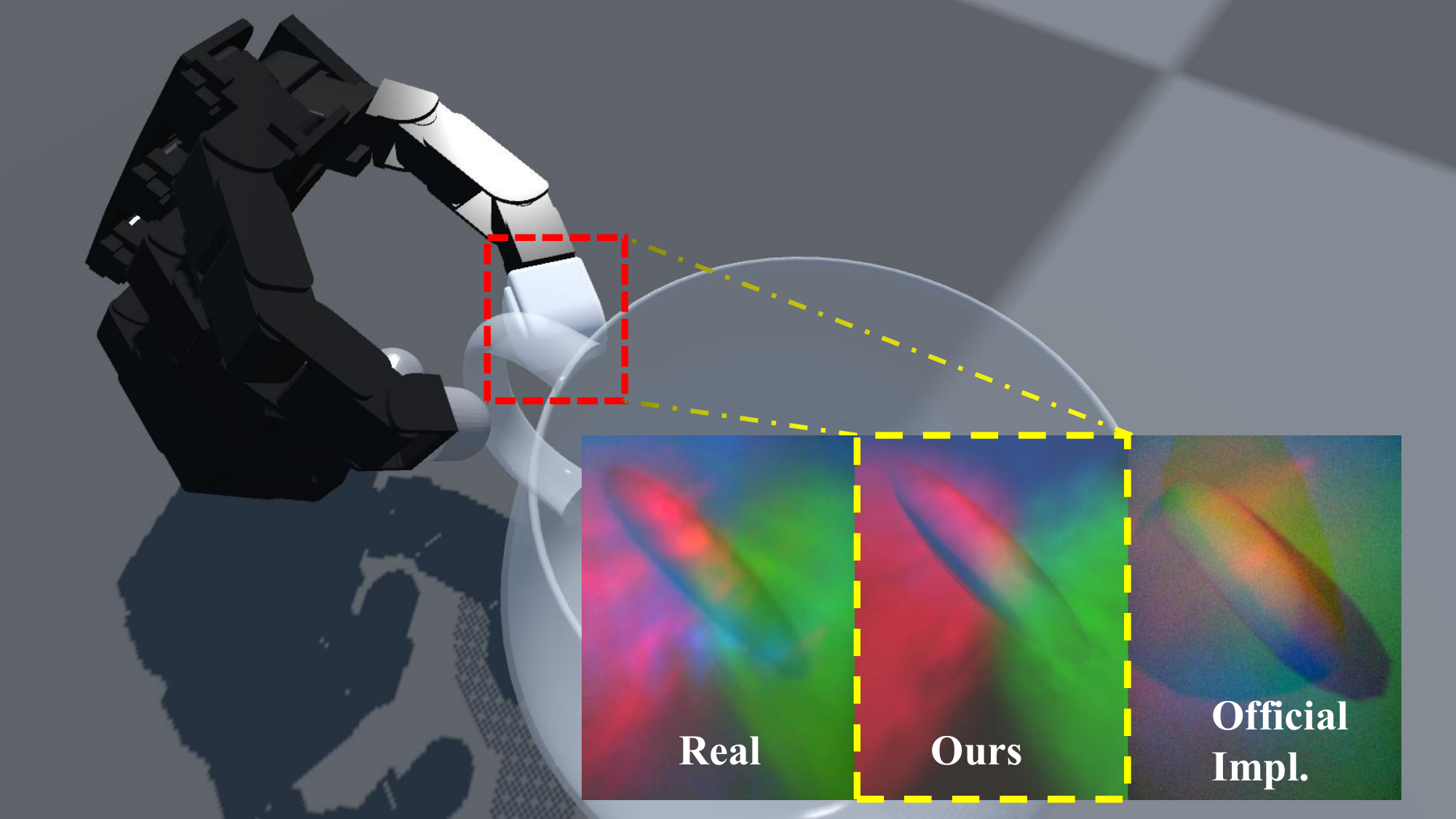}
        \end{minipage}
        \label{digit}
    }
    \subfigure[Force sensor.]{
        \begin{minipage}[t]{0.45\linewidth}
            \includegraphics[width=1\linewidth]{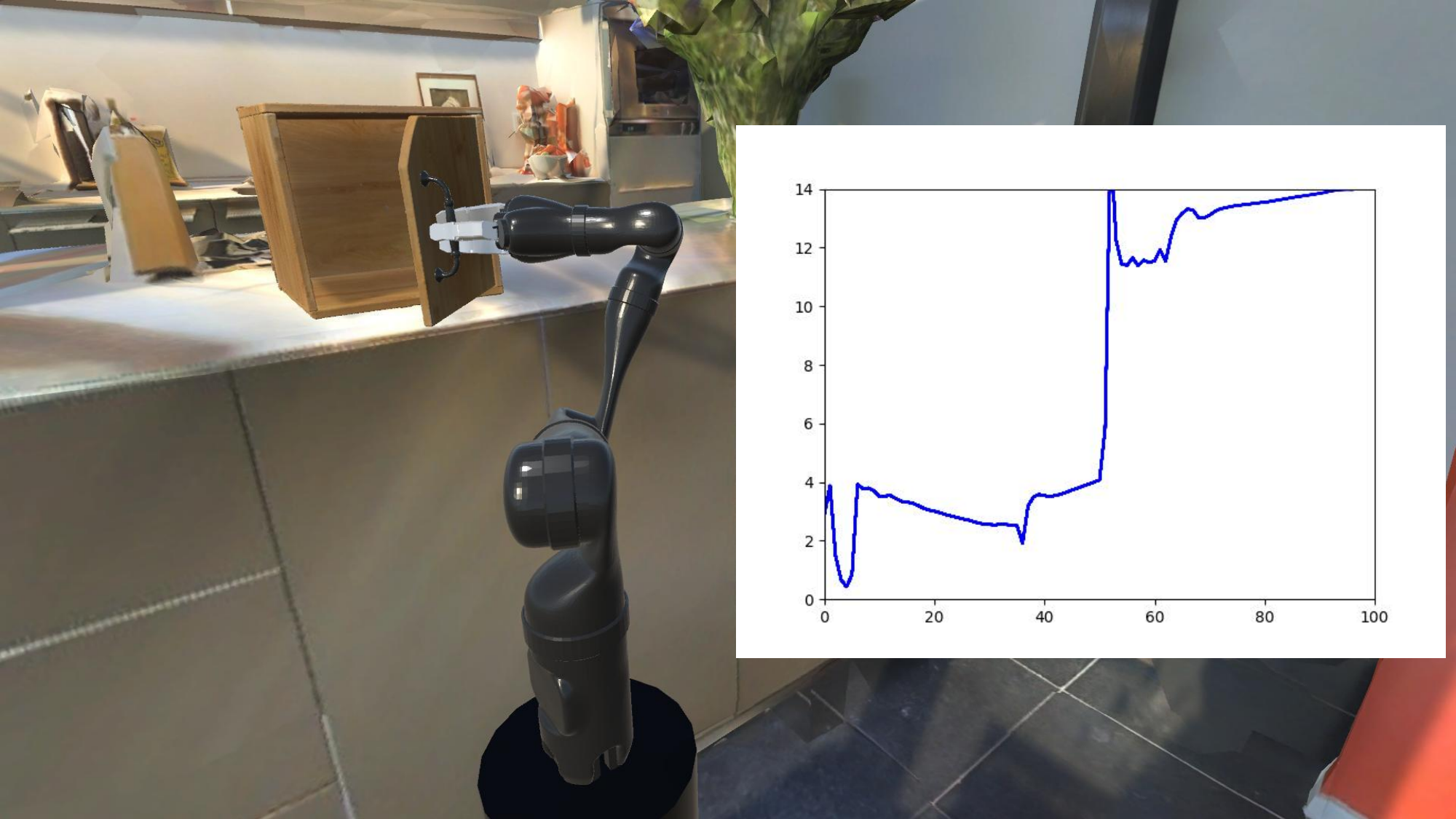}
        \end{minipage}
        \label{force_sensor}
    }
    \caption{Multi-modal Sensor. (a) Visual sensor: RGB, Instance mask, Perfect depth and Normal map; (b) IR-based depth sensor; (c) DIGIT Tactile Sensor with tactile image. We also compare our implementation with real-world DIGIT sensor and official implementation in Pybullet to show that the gap between DIGIT in  real-world in RFUniverse is significantly smaller. For the real-world one, we take the image in the same setting.; (d) Force/Torque Sensor.}
    \label{fig:multi_sensing}
\end{figure*}

\subsection{Multi-modal Sensing}
To interact with the environment, the agent needs to have multi-modal sensory information. RFUniverse supports a set of physics realistic sensors to equip the agent with the capability of perception. For visual input, we leverage ray-tracing techniques and integrated the IR-based depth rendering proposed in Sapien \cite{sapien2} to RFUniverse, which mimics the sensor noise of IR-based depth sensors like RealSense D415 camera. Besides, we also notice the trending of vision-based tactile sensing research in the robotics community \cite{tac1,tac2,tac3,tac4,tac5}. We re-implement and improve the DIGIT sensing \cite{digit} from TACTO \cite{tacto}, and provide a simulated vision-based DIGIT tactile sensing with a less sim-to-real gap. In Fig. \ref{digit}, we place a mug which has a real-world copy into the scene. We then can compare our implementation of DIGIT sensing in RFUniverse with the original ones in TACTO. For force sensing, RFUniverse also has force/torque sensors in the robot joints as shown in Fig.~\ref{force_sensor}.

\subsection{Assets}
Our assets include objects and agents. The simulation environment supports a wide range of object types. For example, we have \textbf{.obj}, \textbf{.ply}, \textbf{.stl} file format for rigid and softbody object, \textbf{URDF} and \textbf{FBX} for articulated object, \textbf{vtk} for volumetric objects used for FEM simulation. We provide a ready-to-use full set of PartNet-Mobility \cite{sapien}, AKB-48 model repository \cite{akb48}, and Google Scanned dataset \cite{google_scanned}. For robots, we support the most common robot arms, such as Franka Emika Panda, and UR5, and mobile robots including Stretch and Fetch, dual-arm mobile manipulator, for instance, PR2. As for the robot hand, we support Robotiq85, Barrett Hand, Allegro, Schunk 5-finger hand, and Shadow hand. If users want to use their custom objects or agents, they can simply load them on the fly via GUI or Python API.

\subsection{Lightweight ROS-free Motion Planner} In many cases, an easy-to-use lightweight planner can help with prototyping and performing experiments. We modify Moveit! \cite{moveit} to a lightweight ROS-free version, RFMove \cite{rfmove}. It can utilize the OMPL library for motion planning, and support obstacle avoidance by synchronizing the simulation scene and planning scene through Python API. We integrate it into the simulation environment to help alleviate the pain of a heavy stack when trying to use libraries for planning.

\subsection{Gym-like Wrapper for Reinforcement Learning} 
As for reinforcement learning, we provide a standard gym-like wrapper for different kinds of environments. It supports multi-thread parallelization, which means users can train multiple agents simultaneously. 
We will later detail the RL experiments in Sec. \ref{sec:exp}, and we leave the basic manipulation experiments in supplementary materials.


\section{Experiments}\label{sec:exp}
To verify the fidelity of RFUniverse, we perform three experiments involving multiphysics coupling effects for hydrodynamics, aerodynamics, and thermodynamics in Sec. \ref{sec:parallelexp} and compare them with the ones in the real world. We perform reinforcement learning for manipulation tasks involving multiphysics coupling effects in Sec. \ref{sec:rl}. We showcase the usage of our lightweight planner for melted butter manipulation task in Sec. \ref{sec:planning}. We also perform experiments for visual pre-trained encoders with reinforcement learning. 
\subsection{Machine Specification}

During training, all experiments are benchmarked on Ubuntu 22.04 platform with Intel(R) Core(TM) i9-10900K CPU and 1 NVIDIA Geforce GTX 1080Ti graphic card. Due to different calculating burdens in the tasks, we dynamically adjust the action time step. We set the action time step for rigid object interaction $t_a^r=\frac{1}{50}s$ , for deformable object, fluid, and air interaction $t_a^d=\frac{1}{25}s$, and for cutting task $t_a^c=\frac{1}{20}s$. 

\subsection{Physics Simulation Verification Experiments}\label{sec:parallelexp}

\begin{figure}[th!]
    \centering
    \includegraphics[width=0.45\textwidth]{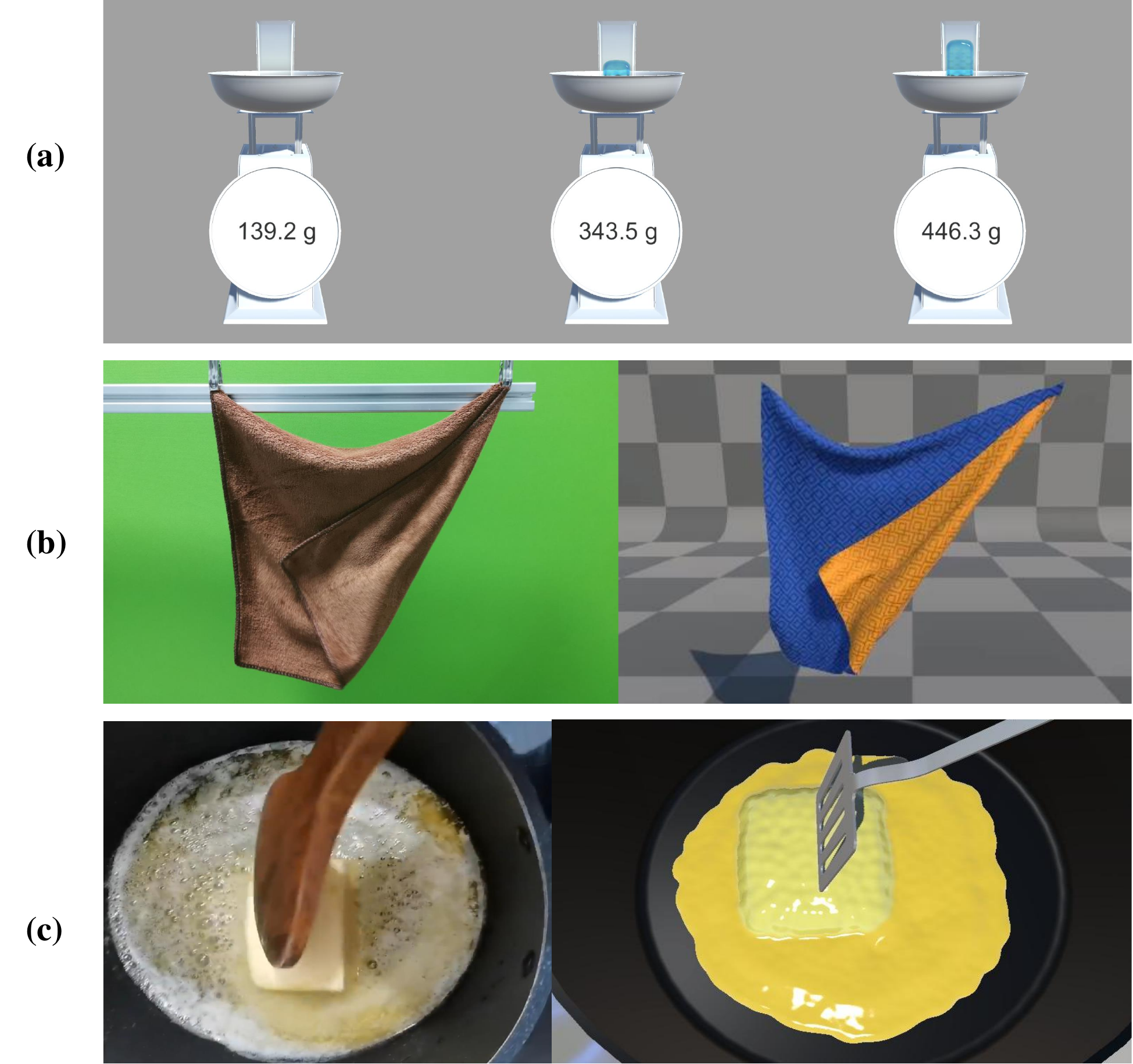}
    \caption{Physics simulation verification experiment setting. (a) depicts water force in RFUniverse. Water is modeled with particles and each particle weighs about 1$g$. The left image contains an empty glass that weighs around 140$g$, while the glass in the middle image contains 200 particles and the glass in the right image contains 300 particles; (b) contains two cloth swaying in the wind in the real world and RFUniverse, which present a similar configuration; (c) contains two butter-melting scenes in the real world and RFUniverse.}
    \label{fig:parall_experiments}
\end{figure}

\paragraph{Water Weight}  We place a cup onto a weighing scale as shown in Fig. \ref{fig:parall_experiments} (a). Then, we slowly add the water to the cup. Since the volume ($V_w$) and mass of water  ($m_w$) satisfy $m_w=\rho \cdot V_w$, where $\rho$ is considered as density and is calibrated from real-world, we can verify the weight reading by checking the volume of the water in the cup.

\paragraph{Towel in Wind} In the real world, we fix a towel to a pole and set up a hair dryer with different levels of wind aside. We record a video of it, and set up the same setting with similar wind conditions. We observe the video and find out they have similar trends. 

\paragraph{Heat Transfer} We take a video clip of a human pushing a slice of butter on a heated skillet and imitate the pushing behavior from the video in simulation. We can observe the simulated butter is melted down in a similar way. 

\begin{figure}[h!]
    \centering
    \includegraphics[width=0.5\textwidth]{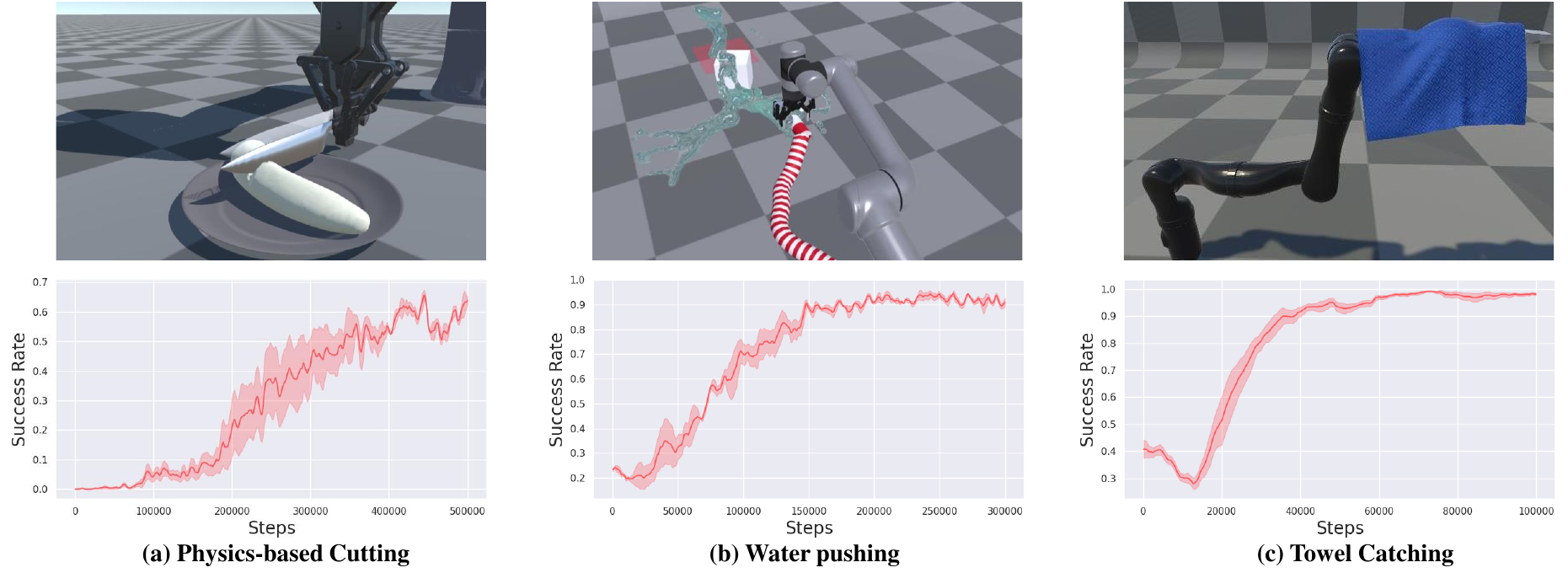}
    \vspace{-5mm}
    \caption{Experimental results for manipulation experiments with reinforcement learning. \textbf{Upper:} The snapshots of task scenes. \textbf{Bottom:} The relationship between success rate and training step for manipulation tasks are displayed.}
    \label{fig:experiments}
\end{figure}

\subsection{Manipulation experiments with Reinforcement Learning}\label{sec:rl}
\subsubsection{Physics-based Cutting}
\paragraph{Task Description} A banana is placed on the ground, and a Flexiv robot arm with an AG-95 gripper holds a knife in its hand. In each episode, the robot will reset to a fixed pose and the task is divided into two phases. In the first phase, a goal pose of the knife is given and the Flexiv robot arm should move and make the knife-in-hand closer to the given pose. This pose is sampled on the top surface of the banana with proper orientation. Then in the second phase, the Flexiv robot will move along its end-effector's Z-axis to cut the banana.
\paragraph{Task Implementation} In each step, a 6-d action is needed with the first 3 dimensions representing the delta translation of the gripper along the X, Y, and Z axis and the last 3 dimensions representing delta Euler angles along X, Y, and Z axis. The observation includes the pose and velocity of the end-effector, the pose of knife-in-hand, and the goal knife pose. All orientations in observation are in quaternion format. We use a two-stage sparse reward function in this task: the default reward is 0, and if the distance between knife-in-hand and goal knife is less than 4$cm$, the reward will add 0.5; if the minimal angle between knife-in-hand and goal knife is less than 5 degree, the reward will add 0.5. The task is regarded as successful when the reward is 1. Since this is a goal-conditioned RL task, we use SAC \cite{sac} algorithm with HER \cite{her} replay buffer to train the policy for this task.

\subsubsection{Water Pushing}
\paragraph{Task Description} A fixed-base UR5 robot arm is placed in the scene, with a Robotiq85 gripper as its end-effector. A 10$cm$ cube is placed outside the reachable space of the UR5 robot. A water pipe is grasped by a Robotiq85 gripper and water is flowing out from the pipe. The robot controls the orientation of the pipe to make the cube move to a target area (red area in Fig. \ref{fig:experiments}(b)) with water. It is similar to the pushing task, so we call it water-pushing.
\paragraph{Task Implementation} In each step, a 4-d action is needed with the first 3 dimensions representing the delta translation of Robotiq85 along the X, Y, and Z axis and the last 1 dimension representing the gripper width. Note that the initial velocity and volume per unit time of water spray are proportional to gripper width. In each episode, UR5 is reset to a fixed pose and the cube will be sampled within a range. The target area will be sampled within a range around the initial cube position. The observation includes the position, velocity, and width of the gripper and cube, as well as the center of the target area. The reward function is the negative distance between the cube and the center of the target area. The task is regarded as successful when the distance between the cube and the target area center is less than 10$cm$. This task is also goal-conditioned, so we use the SAC algorithm with HER.

\subsubsection{Towel Catching}
\paragraph{Task Description} A Kinova Gen2 robot arm is fixed in the scene and a towel will fall down. When the towel is falling down, a wind with random strength and orientation is added to the scene to add some randomization to the movement of the towel. The movement of the end-effector is restricted to a horizontal plane with a fixed height lower than the initial falling height of the towel, so it is required to predict the falling point and catch the towel.

\paragraph{Task Implementation} In each step, a 2-d action representing the delta movement along the fixed horizontal plane is needed. In each episode, the Kinova Gen2 robot arm will be reset to a fixed pose but the towel will fall from a random position with a wind of random strength and orientation added to the scene. Since the towel is a high-dimensional deformable object, we calculate the average position among all vertices of the towel. Thus, the observation includes the position and velocity of the end-effector, the average position and velocity of the towel and initial falling position, wind orientation, and strength. The reward function is shown as Equ. \ref{towel_reward}:
\begin{equation}\label{towel_reward}
    reward=2 - tanh(10\cdot |v_c^Y|)-tanh(10\cdot |\mathbf{v_e}|),
\end{equation}
where $v_c^Y$ is the Y-axis velocity of cloth and $\mathbf{v_e}$ is the 3-d velocity of the end-effector.
The task is regarded as successful if the towel is caught at the end of an episode. The task is trained with the SAC algorithm.

\subsection{Butter Pushing: Manipulation experiments with Planning \& Control}\label{sec:planning}
\paragraph{Task Description} By taking the reference video of butter melting, we first mark the trajectory waypoints in RFUniverse. A spatula is attached to the end-effector of a Flexiv robot arm and pushes the butter to follow the trajectory. 
\paragraph{Task Implementation} The inverse kinematics calculation and the continuous trajectory are produced by RFMove \cite{rfmove}. To make the butter melt similarly to the reference video, we use 200 particles to simulate the butter from solid to liquid. In the beginning, all particles are solidified to form a butter cube, \ie all particles are with high viscosity and surface tension value. When particles at the bottom collide with the pot's collider, their viscosity and surface tension value decrease and act like a melted liquid. Let $T_i$ be the temperature of the particle $i$. To simulate the heat transfer process, the simulation follows~\cite{heattrans}, and applies $\frac{dT_i}{dt} = -T_i/D_r$ to the particles, where $D_r$ denotes the radiation half time. The value of $D_r$ can be used to suggest the cooling speed of the particles.
When bottom particles become liquid, particles with a higher initial position will fall and collide with the pot. Meanwhile, in each step, all particles within a fixed margin will average their viscosity and surface tension. The overall process simulates the heat transfer effect of butter melting from the bottom to the top. All parameters in the experiment are hand-crafted to align with the reference video.

\subsection{Visual pre-trained encoder with Reinforcement Learning}\label{sec:pretrain}
\begin{figure}[h!]
    \centering
    \hspace{-2mm}
    \subfigure[Cluttered and structured scene screenshots for Franka Reach task.]{
        \begin{minipage}[t]{0.95\linewidth}
            \includegraphics[width=1\linewidth]{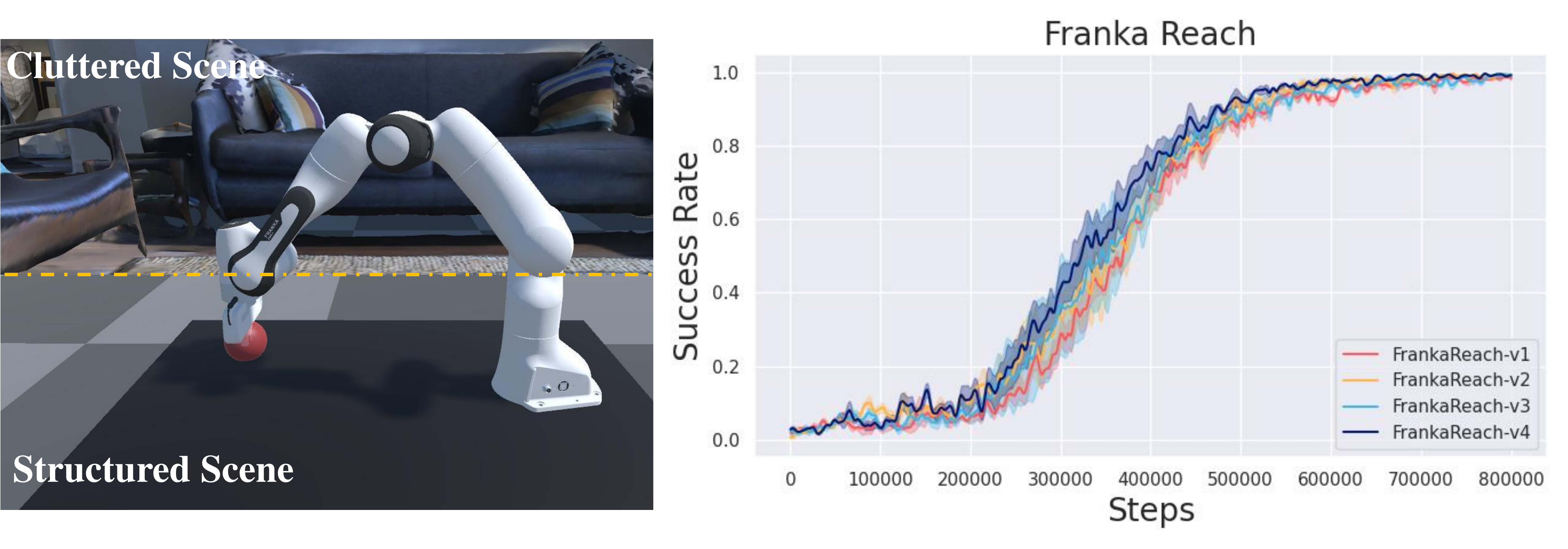}
        \end{minipage}
    }
    \subfigure[Cluttered and structured scene screenshots for Franka Cabinet Closing task.]{
        \begin{minipage}[t]{0.95\linewidth}
            \includegraphics[width=1\linewidth]{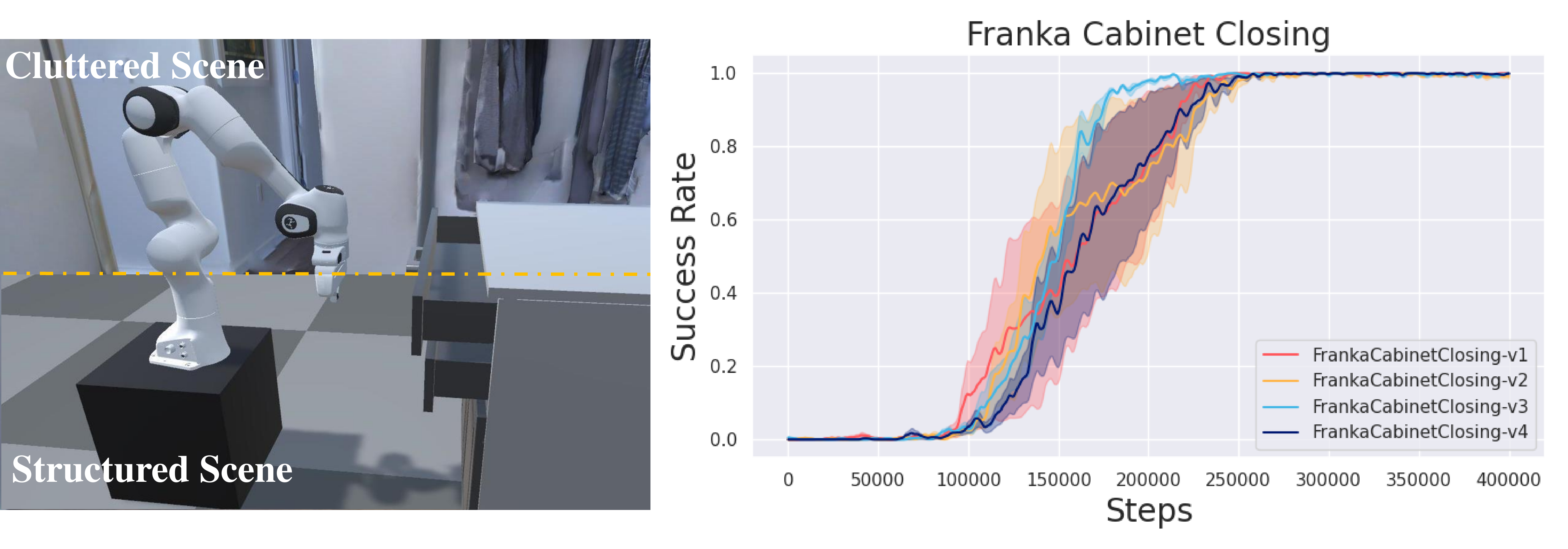}
        \end{minipage}
    }
    \caption{Scene screenshots and experimental results for visual pre-trained encoder with reinforcement learning. Cluttered means with the background of the room environment, while structured means without the background. In legend, \textbf{v1} means eye-on-hand with structured scene; \textbf{v2} means eye-on-base with structured scene; \textbf{v3} menas eye-on-hand with cluttered scene; \textbf{v4} means eye-on-base with cluttered scene.}
    \label{fig:visual_pretrain}
\end{figure}

In these experiments, we follow the setting in \cite{mvp} where we use a pre-trained encoder to encode the image from the current camera perspective. Note that the vector after encoding is the only observation of RL algorithms and the weights of the pre-trained encoder are fixed during training. We build two tasks named Franka Reach and Cabinet Closing and use the identical encoder. For each task, we provide a structured scene version and a cluttered scene version, as well as a camera from a first-person perspective (eye-on-hand) and third-person perspective (eye-on-base) for each version respectively. We use PPO \cite{ppo} algorithm to train both versions with both perspectives for each task. The experimental results for the relationship between training steps and success rate are shown in Fig. \ref{fig:visual_pretrain}.

\subsubsection{Franka Reach} Franka robot resets to a fixed position $\mathbf{p}_f$ and a target position is randomly sampled from the $30cm\times 30cm\times 30cm$ space around $\mathbf{p}_f$. Franka robot arm needs to let its gripper's grasp point reach the target position. The tolerance is 5$cm$ and the target range is a sphere with 5$cm$ radius, which is highlighted in the scene. The reward is the negative distance between the grasp point and the target position.

\subsubsection{Franka Cabinet Closing} A double-layer cabinet with prismatic drawers is initialized with two drawers randomly opened and in a random pose. The Franka robot arm needs to push two drawers back. The reward is the sum of the negative open length of drawers.

\section{Conclusion and Future Work} 
In this paper, we present a novel multiphysics-based simulation environment named \textbf{RFUniverse}. With user-friendly interfaces and a Unity-based backend, it provides a chance for the users to seamlessly interact with the virtual world which supports aerodynamics, hydrodynamics, and thermodynamics. We aim at taking the first step towards building a physics realistic world that supports multiphysics coupling effects, with the hope of extending the task scope of current robot manipulation in simulation to the next level. 


We would like to propose such a question: Do intelligent agents really need high-precision simulation for embodied AI? As humans, we never compute accurate physics when we are interacting with the real world. Instead, we observe and learn to react. Therefore, we can assume that the world human (or any sensors) observe is also noisy. Seemingly realistic physics simulation could already support many tasks. And for policy learning, such computational inaccuracy could bring more diversity in sensory information and benefit the generalization ability of learned models. Similar ideology is presented in domain randomization techniques.

We regard the RFUniverse simulation environment as an infrastructure for multiphysics-based robot learning and will integrate more physics solvers into the system to keep enlarging the research scope of embodied AI.



\bibliographystyle{plainnat}
\bibliography{main}

\end{document}